\documentclass[10pt,twocolumn,letterpaper]{article}

\usepackage{cvpr}
\usepackage{times}
\usepackage{epsfig}
\usepackage{graphicx}
\usepackage{amsmath}
\usepackage{amssymb}
\usepackage{subfig}

\usepackage[pagebackref=true,breaklinks=true,letterpaper=true,colorlinks,bookmarks=false]{hyperref}

\cvprfinalcopy % *** Uncomment this line for the final submission
\def\cvprPaperID{****} % *** Enter the CVPR Paper ID here

\begin{document}
	
	%%%%%%%%% TITLE
	\title{The role of ego vision in view-invariant action recognition}
	
	\author{Gaurvi Goyal, Nicoletta Noceti, Francesca Odone  \\
		{\small DIBRIS - University of Genova, IT}\\
		{\tt\footnotesize {name}.{surname}@dibris.unige.it}
		\and
		Alessandra Sciutti\\
		{\small Istituto Italiano di Tecnologia} \\
		{\tt\footnotesize alessandra.sciutti@iit.it}
	}
	
	\maketitle
	%\thispagestyle{empty}
	
	%%%%%%%%% ABSTRACT
	\begin{abstract}
		Analysis and interpretation of egocentric video data is becoming more and more important with the increasing availability and use of wearable cameras. Exploring and fully understanding affinities and differences between ego and allo (or third-person) vision is paramount for the design of effective methods to process, analyse and interpret egocentric data. 
In addition, a deeper understanding of ego-vision and its peculiarities may enable new research perspectives in which first person viewpoints can act either as a mean for easily acquiring large amounts of data to be employed in general-purpose recognition systems, and as a challenging test-bed to assess the usability of techniques specifically tailored to deal with allocentric vision on more challenging settings.
Our work, with an eye to cognitive science findings,  leverages transfer learning in Convolutional Neural Networks to demonstrate capabilities and limitations of an implicitly learnt view-invariant representation in the specific case of action recognition.
	\end{abstract}
	%%%%%%%%% BODY TEXT
	\vspace{-15pt}
	\section{Introduction}
	Action recognition is a core topic in computer vision with  applications in a variety of artificial intelligence systems --- there included, human-computer interaction, robotics, video-surveillance, just to name a few. We are experiencing today considerable leaps forward in the action recognition research, with better algorithms and models being proposed more and more frequently. 
	Among the open problems the research community is dealing with, we focus on the tolerance to view-point changes. This property is not easily obtained in recognition tasks, and requires special care. In the domain of ego-vision, view invariant action recognition is an important element with two different implications: first, ego-vision systems may provide us with large amount of data streams, which could be fuelling general purpose recognition systems; conversely, in the design of the algorithms for an ego-vision system, one may want to incorporate information learnt from allocentric vision data. While action recognition from egocentric view has been explored to some degree \cite{nguyen2016recognition,song2015activity}, within view-invariance the subject remains largely untouched.
	Over the years, the problem of view-invariant motion recognition has been addressed considering two different settings, i.e. observing the same dynamic event simultaneously from multiple cameras 	\cite{zheng2012cross,li2012discriminative}
	 or considering independent instances of a same dynamic concept \cite{junejo2011view,huang2012discriminative}.

	Methodologically, early works approached it as an epipolar geometry problem \cite{syeda2001recognizing,yilmaz2005recognizing}, while later works can be categorized into methods acting at a descriptor level, to design  representations explicitly embedding view-invariant information \cite{junejo2011view,li2012cross,huang2012discriminative,nguyen2016recognition},  or at a similarity level  
	\cite{wu2012view,huang2012recognizing,zheng2013learning}.
	In this category are also methods addressing the problem with a transfer learning formulation,   from one view to another \cite{zheng2012cross}, or to a common virtual view, sometimes in a 3D reference frame \cite{li2012discriminative}.
	In the last decade, the availability of affordable 3D acquisitions systems has facilitated approaches combining multiple types of information, e.g. videos and skeletal data (for more details see \cite{singh2019human}).
	\begin{figure}%[t]
		\centering
		\subfloat[View0: lateral]
		{\includegraphics[width=0.29\columnwidth]{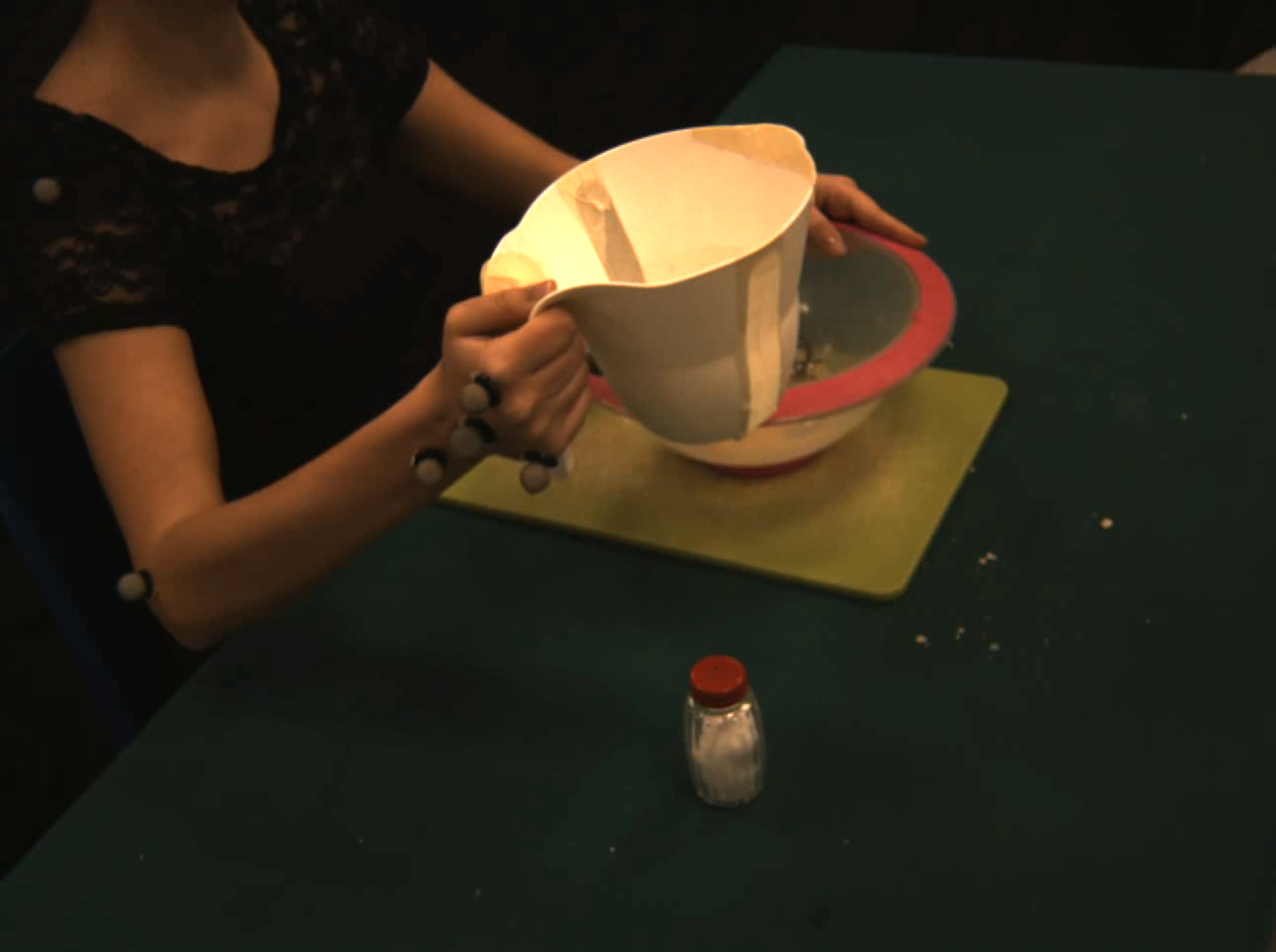}} \,
		\hspace{7pt}
		\subfloat[View1: ego]
		{\includegraphics[width=0.29\columnwidth]{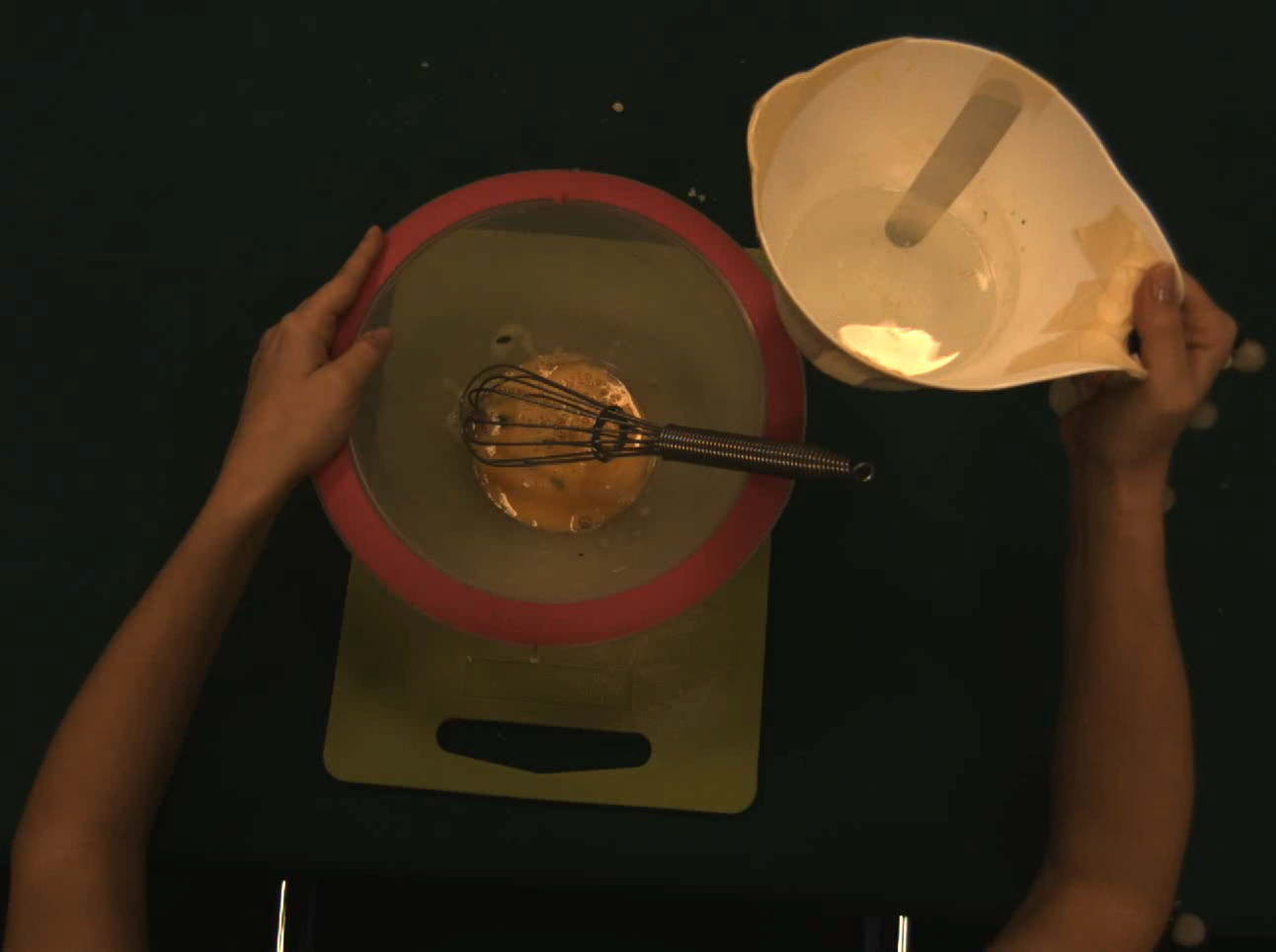}} \,
		\hspace{7pt}
		\subfloat[View2: frontal]
		{\includegraphics[width=0.29\columnwidth]{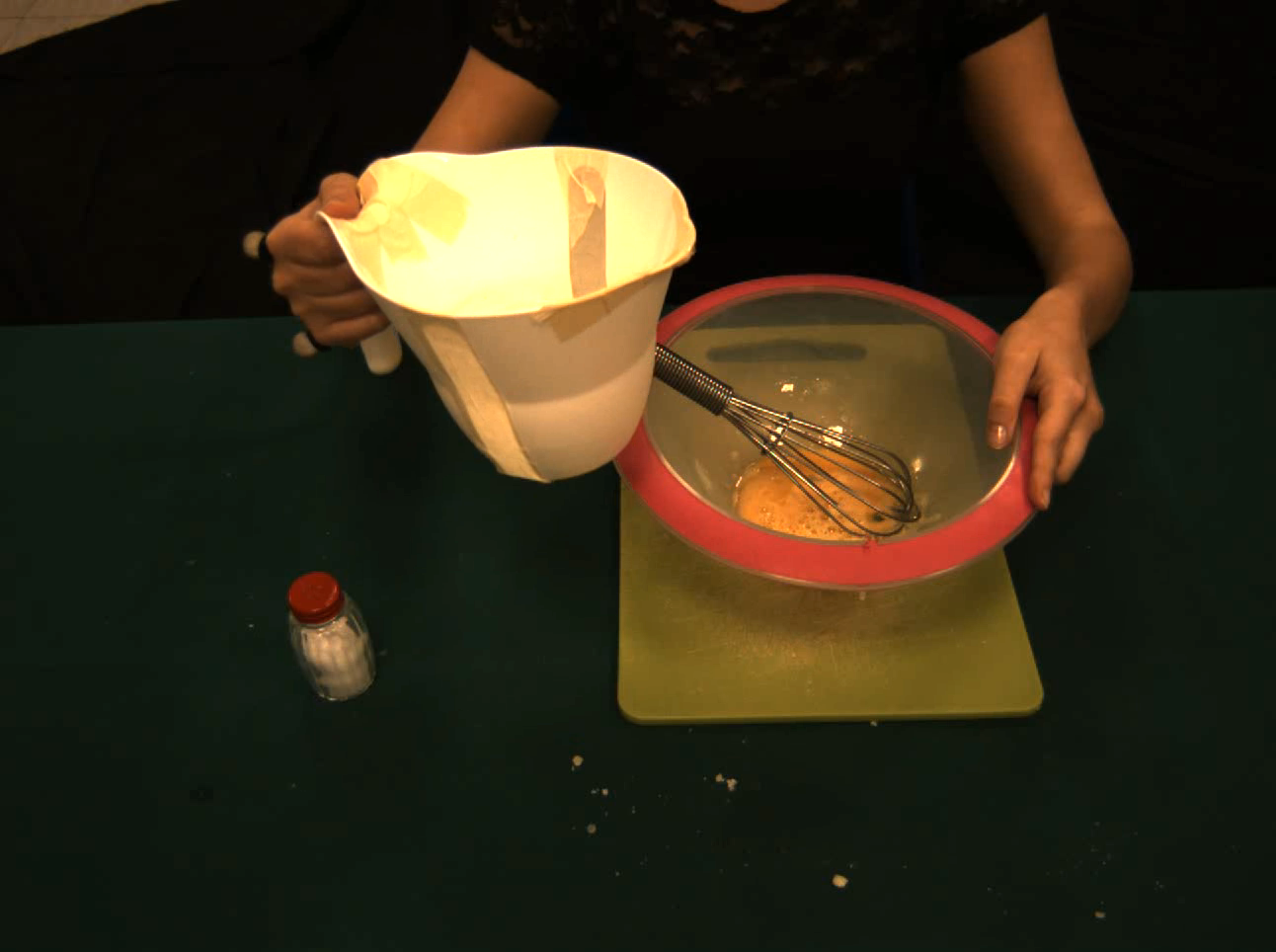}} \,
		%	\hspace{15pt}
		\caption{Synchronized frames from the MoCA dataset (action {\em pouring}).}
		\label{fig::cookingactionsexample}
	\end{figure}
	
View-invariant action recognition plays a crucial role in humans, supporting the capability to solve the correspondence problem, i.e., identifying  a  mapping between the others' actions and their own, which is necessary for crucial activities like social  learning,  imitation or mimicry  \cite{nehaniv2002correspondence}. 
From results in neuroscience, it emerges that such view invariance is a property of higher order visual areas, such as the Superior Temporal Sulcus \cite{grossman2010fmr}, which could however also be supported by pooling together the responses of other view-dependent areas. Indeed, from studies in the macaque brain it is suggested that view-dependent mirror neurons in the premotor cortex (area F5) play an essential role in the formation of view-invariant representations.
Alternatively, it has been speculated that a top-down stream of information from view-dependent mirror neurons might modulate the activity of visual representations in the STS, reinforcing the processing of visual patterns that are associated with different views of the same action \cite{caggiano2011view}. 
Although some human perceptual abilities are immediately available, being part of an innate background of skills, a large portion of them are acquired over time leveraging what we can generally call  the ``human experience''. 
In modern artificial intelligence,  its role is often played by a large amount of data.  
	With the success of data driven methods, especially deep learning, the variety in the available data has a corresponding effect on the variety and the effectiveness of applications. Very complex architectures leverage on the availability of large datasets, which allow us to learn not only input-output relationships with good generalization properties, but also multiple intermediate representations, that could be exploited to address other tasks, through transfer learning. The ability of pre-trained deep neural networks to extract relevant information from new data is documented \cite{Weiss2016} and applied often, but only recently in action recognition \cite{carreira2017,Wang2016}.
	 
Considering this context, in our work we are assessing the potential of pre-trained features in mimicking the role of view-dependent neurons and view invariant higher level descriptions. We consider the MoCA (Multimodal Cooking Actions) \cite{malafronte2017investigating} dataset specifically acquired to study view invariance, both in artificial and biological systems\footnote{The dataset will be soon made available to the research community.}. It includes three different views (an egocentric and two allocentric) of a set of 20 different upper body activities --- see Fig. \ref{fig::cookingactionsexample}. We discuss the effectiveness of intermediate pre-trained features, in dealing with different degrees of view invariance, with specific reference to situations in which the egocentric one is involved. 

	\section{Methodology} \label{sec::methodology}
	{
		Our approach is based on learning intermediate level features with the help of a pre-trained architecture, and applying this representation as an input to a multi-class classification architecture which depends on the specific task of interest. 
		To learn the representation we consider  a variant of the Inception 3D model \cite{carreira2017} taking optical flow estimates as inputs and  3D convolutional filters to incorporate and compress both spatial and temporal information. The model is pre-trained on ImageNet dataset \cite{deng2009imagenet} and on Kinetics-400 \cite{kay2017kinetics}.
		Once trained, the network may be seen as a multi-resolution representation of image sequences.
		
		In order to pin-point an appropriate point of extraction of intermediate features, we identify intermediate layers, which should be producing representations tolerant to view point changes, without being too connected to a specific classification task, hence a point 2 layers before the end was selected. %Specifically, we consider the features computed by Inception 3D after 15 layers of the 17-layer-network 
		(see \cite{carreira2017} for details).
		Thereafter, for a given multi-class classification task, segmented video clips of the actions are used as inputs to the action recognition pipeline. From them, the optical flow is extracted, using the TV-L1 algorithm \cite{zach2007duality}. The optical flow is input into the trained Inception 3D model and the activations or learnt intermediate spatio-temporal features are  then fed to a multi-class classifier.  In Section 3, we will compare results obtained with two different classifiers with different degrees of complexity: Single Layered Perceptron (SLP) and a convolutional neural network (3DConv). The simple SLP allows us to comment on the intrinsic ability of the learnt features to deal with view invariance and with the complexity of ego-vision. The more complex 3DConv, shows the potential of the approach under different challenging classification tasks.}
	
	\begin{figure*}[t]
		\centering
		\subfloat[Cam0 \label{fig:alba0}]
		{\includegraphics[width=0.3\columnwidth]{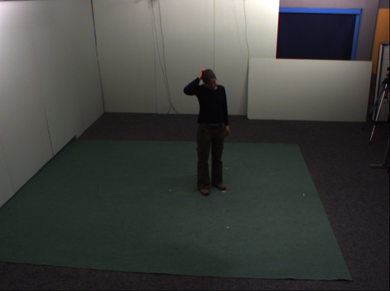}} \,
		\hspace{15pt}
		\subfloat[Cam1 \label{fig:alba1}]
		{\includegraphics[width=0.3\columnwidth]{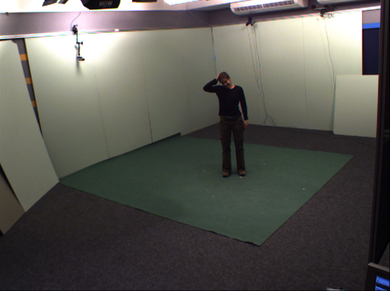}} \,
		\hspace{15pt}
		\subfloat[Cam2 \label{fig:alba2}]
		{\includegraphics[width=0.3\columnwidth]{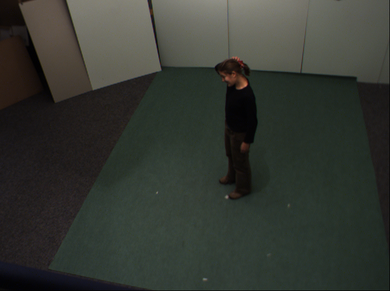}} \,
		\hspace{15pt}
		\subfloat[Cam3 \label{fig:alba3}]
		{\includegraphics[width=0.3\columnwidth]{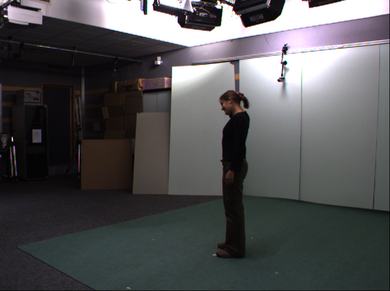}} \,
		\hspace{15pt}
		\subfloat[Cam4 \label{fig:alba4}]
		{\includegraphics[width=0.3\columnwidth]{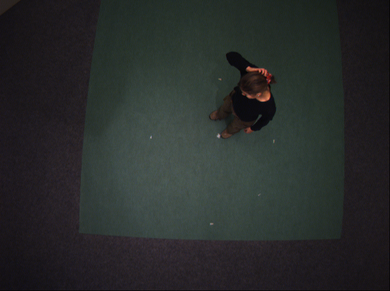}} \,
		\hspace{15pt}	
		\caption{Sample frames from the IXMAS dataset (actor \textit{alba} and action \textit{scratchhead}).}
		\label{fig::ixmasexample}
	\end{figure*}
	
	\begin{table*}[htp]
		\caption{Performance evaluation (in \%) on the MoCA dataset considering various training and test subsets. Views - 0: Lateral, 1: Egocentric, 2: Frontal}
		\begin{center}\vspace{-15pt}
			{
				\begin{tabular}{|c|ccc|c|ccc|cccccc| }
					\hline
					Source$\mid$Target & 0$\mid$0 & 1$\mid$1 & 2$\mid$2 & 0,1,2$\mid$0,1,2 & 0,1$\mid$2 & 0,2$\mid$1 & 1,2$\mid$0 & 0$\mid$1 & 0$\mid$2 & 1$\mid$0 & 1$\mid$2 & 2$\mid$0 & 2$\mid$1 \\
					\hline
					SLP &  93.25 & 91.11 & 92.70 & 87.37 & 68.33 & 46.03 & 68.10 & 47.38 & 68.33 & 47.38 & 32.86 & 66.27 & 34.84  \\
					3DConv &  96.25 & 96.35 & 96.43 & 94.81 & 62.30 & 61.67 & 62.70 & 50.63 & 64.84 & 33.10 & 36.35 & 61.67 & 54.92 \\
					\hline	
				\end{tabular}
			}\vspace{-10pt}
		\end{center}
		\label{tab::cookingresults}
	\end{table*}
	
	\begin{table*}[htp]
		\caption{Performance evaluation on the IXMAS dataset considering a subset of training and test splits of the dataset, based on viewpoints. Mean refers to the average of accuracies for all combinations of viewpoints with one-one protocol. }
		\begin{center}\vspace{-15pt}
			\small{
				\begin{tabular}{|ccccccccccccc| }
					\hline
					Source$\mid$Target & Mean & 0$\mid$1 & 0$\mid$4 &  1$\mid$4 & 2$\mid$4 &  3$\mid$4 & 4$\mid$0 & 4$\mid$1 & 4$\mid$2 & 4$\mid$3  & 0,3$\mid$4  & 0,1,2,3$\mid$4 \\
					\hline
					%				\hline
					DT \cite{wang2011action} & 61.7 & 93.9 & 27.6 & 22.4 & 53.3 & 34.8 & 42.1 & 25.8 & 63.3 & 48.8 & -- & --\\
					Hankelets \cite{li2012cross} & 56.4 & 83.7 & 33.6 & 26.9 & 60.1 & 31.2 &  39.6 & 32.8 & 68.1 & 37.4 & -- & --\\
					SLP & 69.4 & 84.4 & 48.8 & 47.0 & 66.0 & 45.0 & 53.3 & 56.6 & 69.3 & 53.5 & 57.3 & 62.8\\
					3DConv & 68.5 & 89.0 & 44.4 & 42.5 & 61.3 & 45.4 & 48.6 & 49.1 & 57.9 & 46.7 & 49.2 & 57.9\\
					\hline	
				\end{tabular}
			}\vspace{-10pt}
		\end{center}
		\label{tab::ixmasresults}
	\end{table*}
	\begin{figure}%[t]
		\centering
		{\includegraphics[width=0.8\columnwidth]{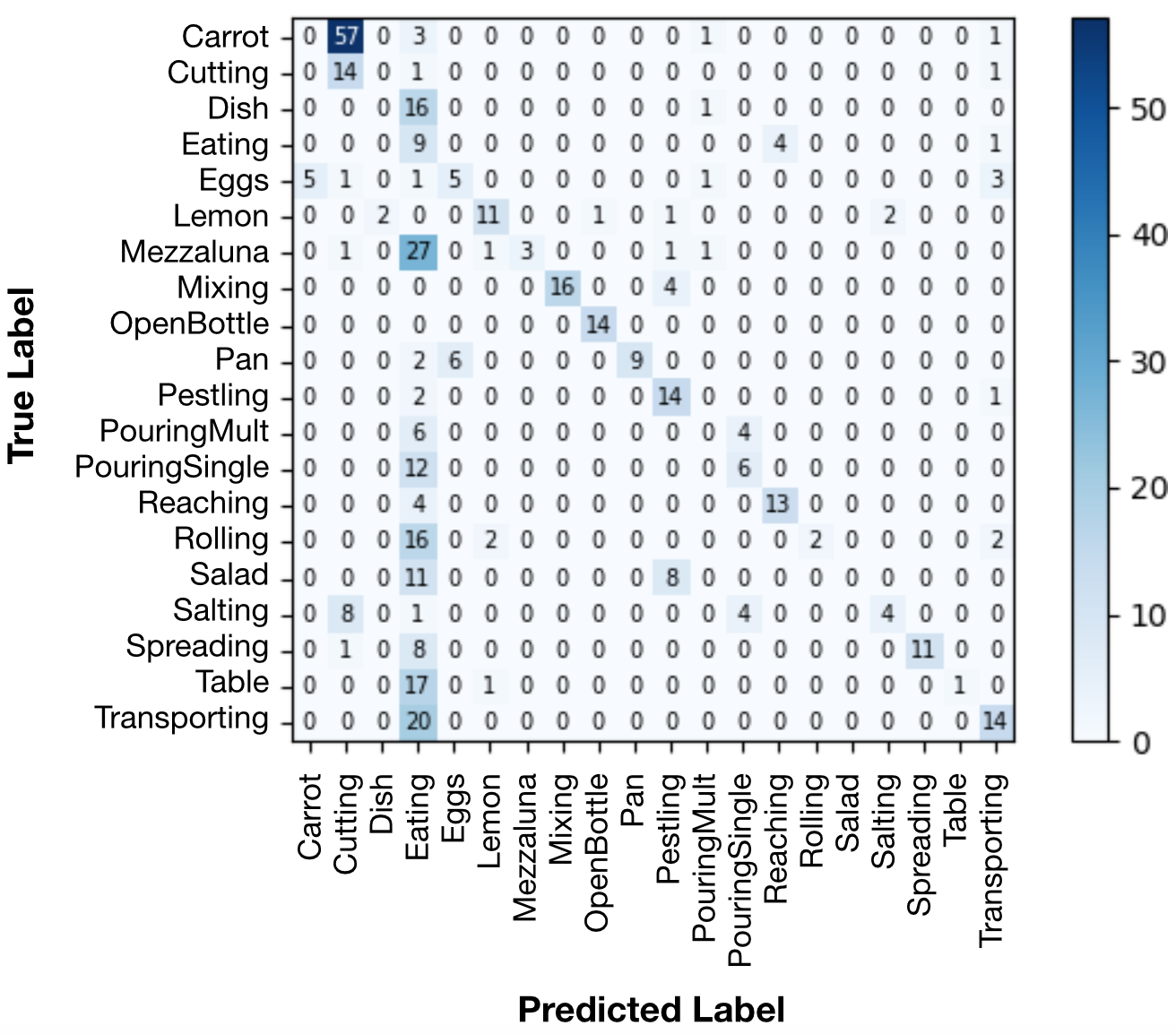}} \, 
		\caption{Confusion matrix for 3DConv classifier  trained on V1 and tested on V2 of the MoCA dataset. }
		\label{fig::confusion1}
	\end{figure}
\section{Results} \label{sec::results}
The core of our experimental analysis focuses on the MoCA dataset \cite{malafronte2017investigating}, consisting of 20 cooking action primitives, involving one or two arms of a volunteer, with subtle differences between different actions. The dataset comprises synchronized videos of actions from 3 different viewpoints see Fig \ref{fig::cookingactionsexample}: Lateral (V0), Egocentric(V1), and Frontal (V2). Training (TR) and Test (TE) sequences are available for each action and viewpoint.
%It is divided into training(TR) and testing(TE) for each action and viewpoint. 
In different iterations of the experiment, we trained the classifiers with a variety of subsets of the TR split and tested on subsections of the TE split. Validation splits were processed using a batch-wise protocol with batch normalization parameters calculated per batch. 
		
	The resulting validation accuracies are shown in Table \ref{tab::cookingresults}.  We carried out a set of baseline experiments, where TR and TE are uniform: (\{i$\mid$i\}, with $i=0,1,2$). We also include another baseline,  (\{0,1,2$\mid$0,1,2\}), where a view-invariant model is obtained simply by training the classifier on multiple views. We can see that in all these cases, the classification performances are high.  Next, we consider a one-view out protocol,  when the classifiers are trained with 2 viewpoints and tested on the third;  in this case, there is a notable and expected drop is the capability of the classifiers to correctly classify the actions, but considering they are not explicitly trained to identify actions view-invariantly, this drop is { modest} and thus not remarkable. Notice in particular how the egocentric view is the hardest to classify if it does not participate in the training phase.
	
	Finally, we adopt a one-one protocol training classifiers on a single viewpoint and evaluating on another viewpoint, to analyse view-view relationship. When both views are allocentric (\{0$\mid$2\},\{2$\mid$0\}),  the resulting values are almost as high as in one-view out experiments. But in all cases where V1 is involved in the one-one protocol (\{0$\mid$1\},\{1$\mid$0\},\{1$\mid$2\},\{2$\mid$1\}), there is a noticeable drop in the performance. The results highlight the specific challenge in dealing with view invariance, when ego-vision is one of the views considered. This appears to be understandable, considering the smaller amount of dynamic information included in the ego view, but it is also in contrast with  findings in cognitive science. 
Indeed, from recent neuro-scientific literature it can be derived that not all views are equally important. First-person view seems to have a prominent role with respect to other perspectives in terms of  responsiveness in the sensorimotor areas of the brain during action observation \cite{angelini2018perspective}
and has been shown to facilitate certain forms of action understanding (e.g., estimating the size of an object to be grasped) \cite{campanella2010visual}.
Beyond egocentric perspective, also the frontal view seems to have a peculiar role, eliciting a stronger activity in the ventral premotor cortex if compared with lateral view, suggesting a preference for ``face-to-face interactions"  \cite{ferri2016stereoscopically}.

		Figure \ref{fig::confusion1} shows the confusion matrix for the above experiment, in case the Conv3D classifier is trained on the egocentric view V1 and tested on V2. Notice that carrot (grating carrots) is almost always classified as cut (cutting a bread). The motion of the two actions is very similar from these two perspectives. Also note, that many actions are often confused with the eating action. This is probably because since the face is not visible in either view, the amount of information available makes it very easy to confuse with actions like transporting (moving an object across the table).  %[COMMENT CONFUSION MATRIX]
	
		We conclude by reporting a further set of experiments, carried out with the same protocol on the IXMAS benchmark (see Fig. \ref{fig::ixmasexample}). This dataset does not include an ego-vision, but incorporates instead a top view which is very different from the others. The results reported in Table \ref{tab::ixmasresults} confirm the observation that the architecture exhibits a good amount of view-invariance, in particular for views that are more likely to be observed. It is instead less robust on view $4$, the top view, which is less common.   Similarly to biological systems, our architecture appears to be better tuned for a set of more likely view points.
	
	\section{Discussion} \label{sec::discussion}
	
Our analysis suggests the relationship between egocentric and individual allocentric viewpoints is significantly less strong than the relationship among allocentric viewpoints (even if they have been acquired by widely different perspectives). 
 This could be explained by the reduced amount of information conveyed by egocentric data,  which is compensated in case of biological vision by proprioception or the awareness of the position and movements of one's own body.%a complementary SELF-AWARENESS?
 
However, the relationship still exists, as is demonstrated by the ability of the classifiers to recognise actions to some extent despite not having any significant information about the egocentric view and how different the actions look from this viewpoint. It is a very interesting observation that the combination of two allocentric viewpoints together were able to train the 3DConv classifier well enough to identify actions from the egocentric viewpoint, with almost the same accuracy as for other scenarios with unseen viewpoints. This apparent ability deserves further investigation to be carried out on wider multi-view datasets, to assess the generality of our observations.
	
\paragraph*{Acknowledgment}
Some results incorporated in this publication have received funding from the European Research Council (ERC) under the European Union's Horizon 2020 research and innovation programme, G.A. No 804388.

	{\small
		\bibliographystyle{ieee}
		\bibliography{references,Bib/ActionRecognitionwithViewInvarience,Bib/ClassicalActionRecognition,Bib/Datasets,Bib/DeepLearning,Bib/DeepLearninginActionRecognition,Bib/MultiViewObjectRecognition,Bib/TemporalActionLocalization,Bib/TransferLearning,Bib/Misc}
	}
\end{document}